%% file: neurips_2026.tex
\documentclass{article}

\PassOptionsToPackage{numbers,sort&compress}{natbib}

\usepackage[main, final]{neurips_2026}

\usepackage[utf8]{inputenc} 
\usepackage[T1]{fontenc}    
\usepackage{hyperref}       
\usepackage{url}            
\usepackage{booktabs}       
\usepackage{amsfonts}       
\usepackage{nicefrac}       
\usepackage{microtype}      
\usepackage{xcolor}         
\usepackage{graphicx}
\usepackage{subfigure}
\usepackage{amsmath}
\usepackage{algorithm}
\usepackage{algorithmic}
\usepackage{nicematrix}
\usepackage{multirow}
\usepackage{booktabs}
\usepackage{tabularx}
\usepackage{nicematrix}
\usepackage{pifont}
\newcommand{\cmark}{\ding{51}}
\newcommand{\xmark}{\ding{55}}

\title{Neuronal Self-Adaptation Enhances Capacity \\ and Robustness of Representation in Spiking Neural Networks}

\author{%
  Zhuobin Yang\textsuperscript{1}\thanks{Equal contribution}\qquad
  Liao Yu\textsuperscript{2}\footnotemark[1]\qquad
  Yeyao Bao\textsuperscript{1}\qquad
  Jian Zhang\textsuperscript{1}\qquad
  Liangfu Lv\textsuperscript{1}\thanks{Corresponding authors}\\
  \bfseries Daqing Guo\textsuperscript{3}\footnotemark[2]\qquad
  Xiaohong Li\textsuperscript{1}\footnotemark[2]\qquad
  Yunliang Zang\textsuperscript{1}\footnotemark[2]\\
  \textsuperscript{1}Academy of Medical Engineering and Translational Medicine, Tianjin University\\
  \textsuperscript{2}School of Mathematical Sciences, Beihang University\\
  \textsuperscript{3}School of Life Science and Technology, University of Electronic Science and Technology of China\\
  \texttt{\{zhuobinyang,yeyaobao\_,jian\_zhang,liangfulv,xhli18,yunliangzang\}@tju.edu.cn} \\
  \texttt{yuliao\_16@buaa.edu.cn} \\
  \texttt{dqguo@uestc.edu.cn}
}

\begin{document}

\maketitle

\begin{abstract}
Spiking Neural Networks (SNNs) are promising for energy-efficient, real-time edge computing, yet their performance is often constrained by the limited adaptability of conventional leaky integrate-and-fire (LIF) neurons. Existing LIF models struggle with restricted information capacity and susceptibility to noise, leading to degraded accuracy and compromised robustness. Inspired by the dynamic self-regulation of biological potassium channels, we propose the Potassium-regulated LIF (KvLIF) neuron model. KvLIF introduces an auxiliary conductance state that integrates membrane potential and spiking history to adaptively modulate neuronal excitability and reset dynamics. This design extends the dynamic response range of neurons to varying input intensities and effectively suppresses noise-induced spikes. We extensively evaluate KvLIF on both static image and neuromorphic datasets, demonstrating consistent improvements in classification accuracy and superior robustness compared to existing LIF models. Our work bridges biological plausibility with computational efficiency, offering a neuron model that enhances SNN performance while maintaining suitability for low-power neuromorphic deployment.
\end{abstract}

\section{Introduction}
Spiking Neural Networks (SNNs) offer biological plausibility, event-driven computation, and strong spatiotemporal processing, with significant energy efficiency on neuromorphic hardware for low-power, real-time applications \cite{ref1, ref2, ref5}. The Leaky Integrate-and-Fire (LIF) neuron model is a core component for its simplicity and biological relevance \cite{ref8}. However, its non-differentiable spiking hinders direct backpropagation, motivating surrogate gradient (SG) methods \cite{ref11, ref12}. Membrane potential reset plays a key role in performance: soft reset preserves residual temporal information, while hard reset discards it \cite{ref11, ref18}. Although soft reset is generally advantageous in preserving temporal information, conventional LIF models still face limitations that constrain their effectiveness in complex temporal tasks.

\begin{figure}[t]
\centering
    \subfigure[Input-output MI across input-intensity ranges.\label{fig1a}]{
        \centering
        \includegraphics[width=0.48\columnwidth]{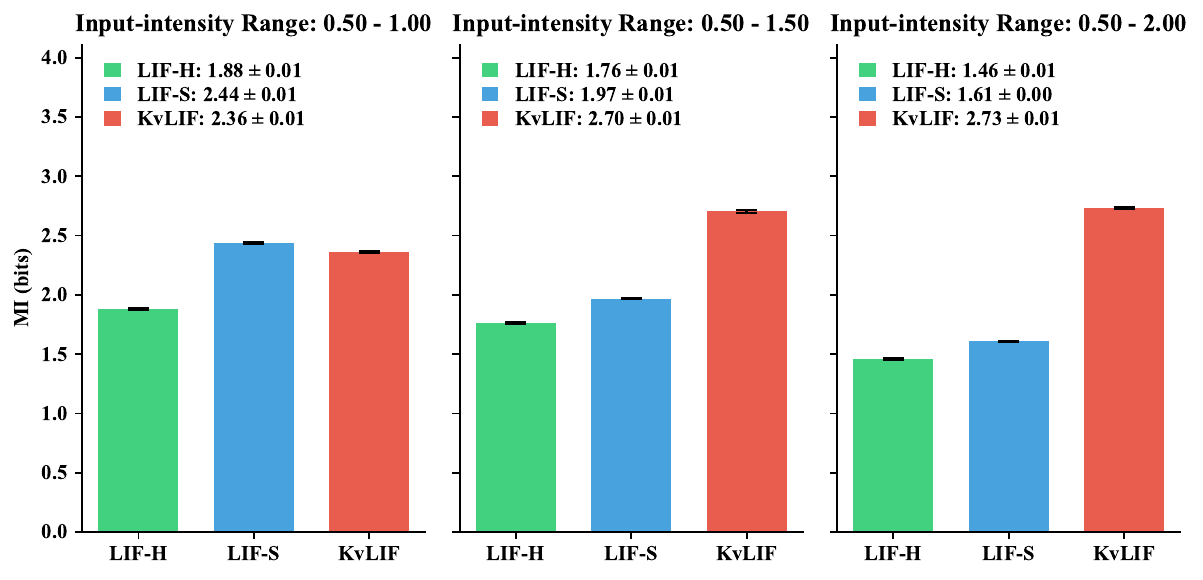}
    }
    \hfill
    \subfigure[Neuronal responses to temporal input. \label{fig1b}]{
        \centering
        \includegraphics[width=0.48\columnwidth]{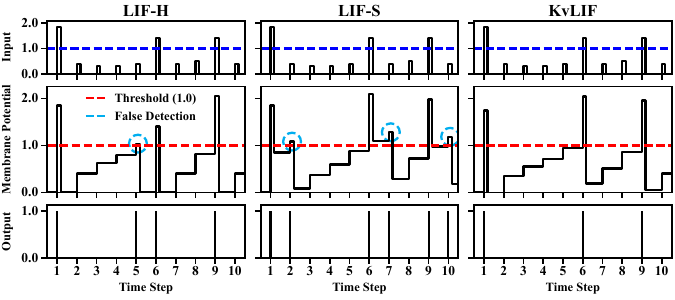}
    }
    \caption{KvLIF enhances dynamic response range and suppresses redundant spikes. LIF-S: LIF model with soft reset; LIF-H: LIF model with hard reset.}
    \label{fig1}
    \vskip -0.1in
\end{figure}

In SNNs, input signals are typically encoded as spike trains whose firing rates increase with input intensity \cite{ref15}. Input intensities in practical settings often span a broad range (see examples in Figure \ref{fig5}, Appendix \ref{a_a}). Neuron models are therefore expected to produce responses that remain discriminable across a wide dynamic range. However, as quantified by input-output mutual information (MI), conventional LIF neurons exhibit limited information capacity at high input intensities (Figure \ref{fig1}(a)). When the upper bound of the input range is increased to 2.00, MI decreases for both hard- and soft-reset variants, consistent with response saturation and diminished information transmission in the high-intensity regime. Conventional LIF neurons also tend to generate redundant spikes under high-intensity inputs because membrane integration continues after salient events (Figure \ref{fig1}(b)). After an event-triggered spike, residual or noise-driven inputs can still accumulate and spuriously elicit additional spikes. This effect is particularly pronounced under soft reset. Together, these observations suggest that conventional LIF dynamics have restricted information capacity and limited excitability control under noisy, high-intensity temporal inputs.

\begin{figure*}[t]
\begin{center}
\centerline{\includegraphics[width=\textwidth]{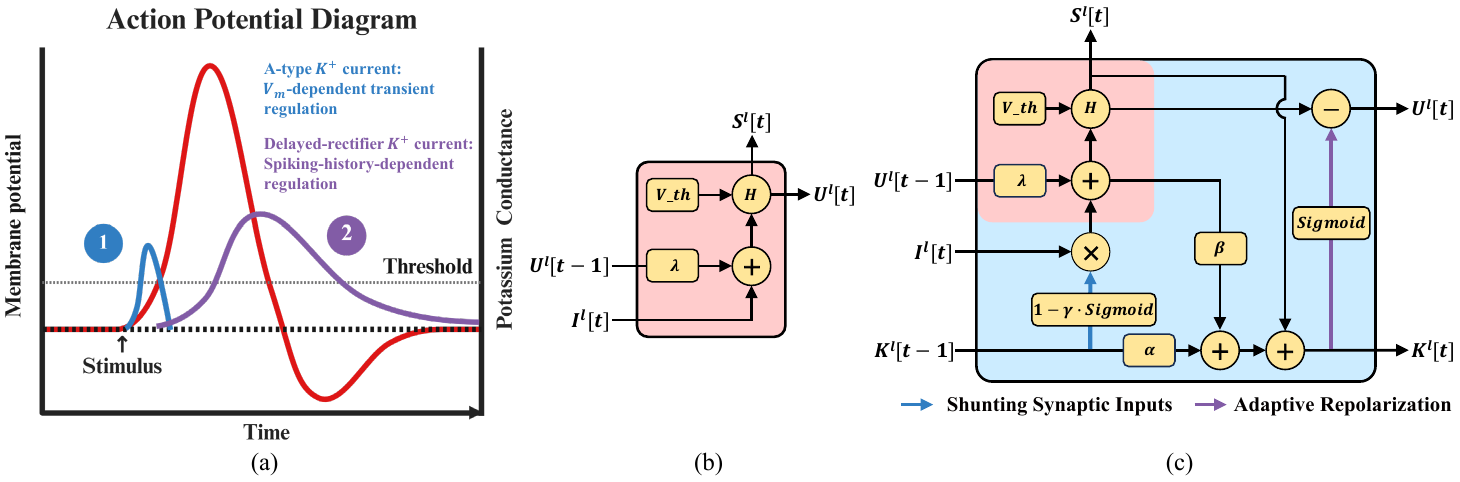}}
\caption{Biophysical motivation and computational realization of potassium-regulated neuronal self-adaptation. (a) A-type and delayed-rectifier potassium currents regulate subthreshold excitability and post-spike repolarization in biological neurons. (b) Conventional LIF performs fixed integration and reset without activity-dependent regulation. (c) KvLIF introduces a potassium conductance state \(K[t]\) to coordinate synaptic shunting and adaptive repolarization, enabling voltage- and spike-history-dependent self-regulation.}
\label{fig2}
\end{center}
\vskip -0.3in
\end{figure*}

Although LIF models resemble neuronal spiking, biological neurons exhibit more dynamic spiking responses due to the influence of diverse biological factors. In particular, neurons contain diverse ion currents that enable self-adaptive behavior under varying input signal intensities \cite{ref17, ref19}. Even within the same type of ion channel, different components can exhibit distinct properties. As shown in Figure \ref{fig2}(a), A-type potassium currents activate within the subthreshold potential range, dampening neuronal excitability to prevent premature action potential generation \cite{ref20}. In contrast, components like delayed-rectifier potassium currents activate only at higher membrane potentials, primarily driving membrane repolarization and contributing to the subsequent afterhyperpolarization \cite{ref21}. These ion channels can modulate their activation and deactivation states based on real-time membrane potential and spiking experience, thereby dynamically adjusting a neuron's response range and contributing to working memory \cite{ref22}.

In this work, inspired by the adaptive neuronal responses, we propose a biologically inspired Potassium Regulated Leaky Integrate-and-Fire (KvLIF) model. By dynamically adjusting its sensitivity range based on spiking history and real-time membrane potential, KvLIF achieves robust spike encoding across a wide range of input intensities in static signals and reduces spurious event detections in temporally rich data (Figure \ref{fig1}). The KvLIF model introduces an auxiliary state variable, \(K[t]\). This variable integrates historical membrane potential fluctuations and spiking activity to adaptively modulate neuronal excitability. 
During the subthreshold phase, \(K[t]\), inspired by A-type potassium channels, imposes voltage-dependent input shunting to suppress membrane depolarization. Upon spike generation, \(K[t]\) drives a dynamic voltage reset mechanism, drawing on the functional role of delayed rectifier potassium channels in biological afterhyperpolarization.
Together, \(K[t]\) incorporates both  membrane potential and spiking experience into the model to adjust neuronal response. This design provides several computational benefits: extended dynamic range, noise robustness, and shorter inference windows, when compared to the conventional LIF models.

The main contributions of this work are as follows:
\begin{itemize}
\item \textbf{Analysis of LIF model limitations}: We provide a detailed investigation of the fundamental limitations of conventional LIF neuron models, including restricted response ranges and vulnerability to noise.
\item \textbf{A biologically inspired adaptive neuron}: 
Motivated by self-regulatory mechanisms in biological neurons, we propose the KvLIF model which features a novel two-stage dynamic regulation mechanism. It first adaptively modulates neuronal states according to membrane potential fluctuations, and then reshapes neuronal excitability based on spiking history. By coupling voltage-driven immediate regulation with spike-history-dependent post-spike regulation, KvLIF achieves a more flexible response profile and improved robustness.
\item \textbf{Comprehensive empirical validation}: We perform extensive experiments across both static and neuromorphic datasets, demonstrating that the proposed KvLIF model consistently surpasses existing neuron models in classification accuracy and robustness to noise.
\end{itemize}


\section{Related Work}
\subsection{Parametric Optimizations of Spiking Neurons}
Conventional LIF neurons are computationally efficient but limited in modeling complex spatiotemporal dynamics. Optimizing neuronal internal parameters has therefore become an effective way to process multi-scale temporal information. For instance, PLIF introduces learnable membrane time constants to adapt integration dynamics through gradient-based training \cite{ref23}, while LTMD employs learnable thresholds to regulate neuronal firing sensitivity \cite{ref24}. Beyond single-parameter optimization, expanding the structured parameter space further improves neuronal expressiveness. GLIF incorporates learnable gating factors to integrate diverse biological features and control information flow \cite{ref25}, whereas TC-LIF adopts a two-compartment architecture with coupling parameters to better capture long-term temporal dependencies \cite{ref26}.

Despite enhanced representational capacity, these parametric methods produce neuronal dynamics that remain static during inference \cite{ref27}, limiting adaptability and robustness in complex dynamic environments.

\subsection{Adaptation Mechanisms in Spiking Neurons}
Various dynamic adaptation mechanisms have proven effective in enhancing network representational capability and gradient propagation efficiency. Spike-frequency adaptation (SFA) LIF neurons, such as ALIF \cite{refr1} and adLIF \cite{refr2}, introduce an additional adaptation variable to encode temporal context and enhance temporal processing. 
To alleviate gradient vanishing, Huang et al. proposed the CLIF neuron model \cite{ref28}, which adds a complementary membrane potential to provide extra backpropagation paths and preserve temporal gradients. Similarly, ILIF introduces two independent inhibitory variables to adaptively balance gradient preservation and neuronal overactivation \cite{ref29}. From the perspective of information retention, Huang et al. designed AR-LIF, which combines adaptive reset with dynamic thresholds to reduce information loss \cite{ref30}. To capture more complex spatiotemporal dependencies, Wang et al. proposed STC-LIF, where learnable autaptic circuits regulate current inputs and historical states to improve predictive capability \cite{ref27}.

Current adaptive neuron models mainly rely on independent, single-stage adaptation \cite{refr1, refr2, ref28}. For example, ALIF implements spike-frequency adaptation with a spike-history-dependent threshold that rises after spikes and suppresses firing. KvLIF instead introduces a potassium-conductance-like state \(K[t]\), driven by both membrane-potential fluctuations and spike history, that shapes both membrane integration and reset. By coupling subthreshold voltage dependence with post-spike repolarization, KvLIF enables more flexible and robust neuronal dynamics, thereby achieving superior performance over existing adaptive neuron models.

\section{Preliminary}
\subsection{LIF Neuron Model}
The LIF model serves as the foundational computational unit in SNNs. As shown in Figure \ref{fig2}(b), the LIF neuron maintains an internal membrane potential that evolves over time. The subthreshold dynamics for neurons in layer \(l\) at time step \(t\) are governed by the accumulation of postsynaptic current \(I^{l}[t]\) and the decay of their previous membrane potential \(U^{l}[t-1]\). The membrane potential update is expressed as:
\begin{equation}
I^{l}[t] = W^{l}S^{l-1}[t] \label{eq1}
\end{equation}
\begin{equation}
U^{l}[t]=\lambda U^{l}[t-1] + I^{l}[t] \label{eq2}
\end{equation}

Here, $\lambda \in (0,1)$ is the membrane decay constant, and \(W^{l}\) denotes the synaptic weights from layer \(l-1\) to layer \(l\). \(S^{l-1}[t]\) represents the spikes of layer \(l-1\)  neurons at time step \(t\). 

Nonlinearity is introduced via a spike generation mechanism. The binary output \(S^{l}[t]\) is generated by the Heaviside function \(H(\cdot)\) when the accumulated potential \(U^{l}[t]\) exceeds the firing threshold \(V_{th}\). The mathematical expression is shown in Eq.\eqref{eq3}:
\begin{equation}
S^{l}[t] = H(U^{l}[t]-V_{th})=
\begin{cases}
1, & U^{l}[t] \geq V_{th} \\
0, & otherwise
\end{cases} \label{eq3}
\end{equation}

After spiking, neurons reset their membrane potentials to prepare for the next firing. This process is illustrated in Eq.\eqref{eq4}:
\begin{equation}
U^{l}[t]=
\begin{cases}
U^{l}[t]-V_{th}S^{l}[t], & soft\quad reset \\
U^{l}[t]\odot(1-S^{l}[t]), & hard\quad reset
\end{cases} \label{eq4}
\end{equation}

Here, \(\odot\) denotes the element-wise multiplication.

\subsection{SNN Training Framework}
The training process relies on the Backpropagation Through Time (BPTT) algorithm \cite{ref45}. To capture temporal dynamics, the gradients for the synaptic weights \(W^{l}\) are calculated by accumulating the derivatives of the loss function \(L\) across all time steps \(T\):
\begin{equation}
\nabla_{W^{l}} \ L = \sum_{t=1}^{T} \frac{\partial \ L }{\partial U^{l}[t]} \frac{\partial U^{l}[t]}{\partial W^{l}}
\end{equation}
To facilitate the backward flow of gradients through the non-differentiable spiking mechanism, the framework incorporates the SG strategy. In this work, the derivative of the spike generation function is approximated using a rectangular function \(\mathbb{H}(\cdot)\):
\begin{equation}
\frac{\partial S^{l}[t]}{\partial U^{l}[t]} \approx \mathbb{H}(U^{l}[t]) = \frac{1}{\sigma} \cdot \mathbb{I}\left( \left| U^{l}[t] - V_{th} \right| < \frac{\sigma}{2} \right)
\end{equation}
where \(\mathbb{I}(\cdot)\) denotes the indicator function, and \(\sigma\) serves as a hyperparameter typically set to the threshold value \(V_{th}\).

\subsection{Voltage-gated Potassium Currents}
Voltage-gated potassium channels are key intrinsic regulators of neuronal excitability. In particular, the A-type potassium current is recruited during the subthreshold depolarization phase. It generates an outward current that shunts excitatory inputs to delay spike initiation and regulate firing frequency \cite{ref31}. After spiking, the delayed rectifier potassium channels dominate the repolarization phase. This current facilitates the restoration of the resting membrane potential and determines the depth of afterhyperpolarization \cite{ref20}. Such coordinated dynamic regulation prevents stochastic firing and is critical for maintaining system stability \cite{ref32}.

\section{Methodology}
\label{sec_met}
\subsection{Limitation of LIF Neuron Model}
For the LIF neuron model with soft reset, when a neuron fires at time step \(t-1\) with a membrane potential \(U[t-1]\) exceeding the threshold \(V_{th}\), the residual potential \(\delta[t-1]\) retained after the soft reset is calculated as:
\begin{equation}
\delta[t-1] = U[t-1] - V_{th}
\end{equation}
where $\delta[t-1] > 0$. According to Eq.\eqref{eq2}, the condition for generating a consecutive spike at time step \(t\) can be rewritten as:
\begin{equation}
\lambda \delta[t-1] + I[t] > V_{th}
\end{equation}
Solving for the input current \(I[t]\), we derive the minimum input current required to generate a spike, denoted as \(I_{min}[t]\):
\begin{equation}
I_{min}[t] = V_{th} - \lambda \delta[t-1]
\end{equation}
Given $0 < \lambda < 1$ and $\delta[t-1] > 0$,  it strictly holds that $I_{min}[t] < V_{th}$. This inequality mathematically proves that the retention of \(\delta[t-1]\) lowers the firing threshold, rendering the model more sensitive to subsequent inputs.

Furthermore, this inertia effect is exacerbated during continuous firing sequences. Consider a scenario where the neuron fires continuously over time steps $t = 1, \ldots, T$, driven by a constant input overload $\Delta I = I - V_{th} > 0$. The residual potential \(\delta[T]\) accumulates recursively. By expanding the recursion $ \delta[T] = \lambda \delta[T-1] + \Delta I$ from an initial state $ \delta[0] = 0 $, we obtain a geometric series:
\begin{equation}
\delta[T] = \Delta I \sum_{i=0}^{T-1} \lambda^i = \Delta I \frac{1 - \lambda^T}{1 - \lambda}
\end{equation}
We now examine the minimum input required for the next spike at $t = T + 1$, which is $ I_{min} [T+1] = V_{th} - \lambda \delta[T]$. Since $\delta[T]$ increases monotonically with \(T\), the minimum input current required for the next spike at \(t = T + 1\) decreases:
\begin{equation}
I_{min}[T+1] < I_{min}[T]
\end{equation}
This demonstrates the cumulative inertia inherent in models with soft reset: continuous spiking monotonically reduces the threshold required to sustain firing. This phenomenon not only progressively diminishes the model's ability to discriminate different input intensities, but also increases its sensitivity to noise.

In the hard-reset LIF model, the membrane potential is forced to zero immediately after spiking. Although this prevents cumulative inertia, it also eliminates the residual potential, hindering the retention of historical activity information and resulting in a limited dynamic response range.

\subsection{The Design of KvLIF Model}
Inspired by the roles of potassium currents in regulating neuronal excitability and post-spike membrane dynamics, we propose KvLIF as a simplified, biologically inspired extension of the LIF model. As shown in Figure \ref{fig2}(c), an auxiliary adaptive state \(K[t]\) integrates membrane-potential and spike-history information to jointly modulate subthreshold integration and post-spike reset. Here, \(K[t]\) serves as a phenomenological abstraction of potassium-mediated feedback, rather than a direct representation of a physical potassium current or conductance.

\textbf{Voltage- and Spike-History-Dependent Adaptive State.} 
For layer \(l\), the adaptive state is first updated using its previous value and the current membrane potential:
\begin{equation}
\hat{K}^{l}[t] = \alpha K^l[t-1] + \beta U^l[t]
\end{equation}
Here, \(\alpha\) controls the retention of the adaptive state from
the previous time step, while \(\beta\) scales the contribution
of the current membrane potential.

Potassium currents exhibit a surge during spiking, reminiscent of delayed rectifier or high-threshold activated potassium currents. This is incorporated by updating \(K^l[t]\) with the spike output \(S^l[t]\):
\begin{equation}
K^l[t] = \hat{K}^{l}[t] + S^l[t]
\end{equation}
The linear coupling of \(K^l[t]\) with \(U^l[t]\) creates a negative feedback loop and serves as the short-term memory to adaptively modulate neuronal excitability before the membrane potential reaches the spike threshold. The coupling of \(K^l[t]\) to \(S^l[t]\) models the explosive opening of potassium channels during spiking, which serves two distinct functions:
\begin{enumerate}
\item \textbf{Rapid repolarization of the membrane potential}, inducing a refractory period immediately after a spike.
\item \textbf{Slow decay of hyperpolarization}, where elevated potassium conductance persists across time steps through the retained fraction \(\alpha\), making \(K^l[t]\) a proxy for long-term spiking history in excitability regulation.
\end{enumerate}

\textbf{Shunting Synaptic Inputs.} In biological neurons, local dendritic potassium channels can shunt postsynaptic responses and reduce excitatory drive. Inspired by this mechanism, we modulate the postsynaptic current \(I^l[t]\) using the potassium channel state variable from the previous time step:
\begin{equation}
I^l[t] = I^l[t] \odot (1 - \gamma\sigma(K^l[t-1]))
\end{equation}
Here, \(\sigma(\cdot)\) denotes the sigmoid function, which maps the potassium current to a normalized activation probability, and \(\gamma\) scales the magnitude of the shunting effect. The term $ (1 - \gamma \cdot \sigma(K^l[t-1])) $ functions as a dynamic shunting gate, decreasing the strength of input currents. This conductance-based gating mechanism regulates neuronal excitability during the subthreshold integration phase, slowing depolarization, preventing excessive spiking.

\textbf{Potassium Channels for Adaptive Repolarization.} To capture somatic repolarization dynamics, we implement an adaptive reset mechanism inspired by biological afterhyperpolarization. In the KvLIF model, the membrane potential is updated as:
\begin{equation}
U^l[t] = U^l[t] - S^l[t] \odot (V_{th} + \sigma(K^l[t]))
\end{equation}
The reset depth here is dynamically modulated by $\sigma(K^l[t])$. Since $K^l[t]$ surges immediately upon spiking, this term forces the membrane potential to drop significantly below the resting membrane potential. Larger $K^l[t]$ values result in deeper resets, effectively extending the relative refractory period. This adaptive reset provides activity-dependent control of subsequent firing.

Building upon these foundational principles, we derive the KvLIF model as:
\begin{equation}
    \begin{aligned}
        U^l[t] &= \lambda U^l[t-1] + I^l[t] \odot (1 - \gamma\sigma(K^l[t-1]))\\
        S^l[t] &= H(U^l[t] - V_{th}) \\
        K^l[t] &= \alpha K^l[t-1] + \beta U^l[t] + S^l[t] \\
        U^l[t] &= U^l[t] - S^l[t] \odot (V_{th} + \sigma(K^l[t]))
    \end{aligned}
\end{equation}

KvLIF coordinates subthreshold integration and post-spike reset through a shared adaptive state driven by membrane potential and spiking history (Algorithm \ref{alg1}, Appendix \ref{a_r1n}). This formulation captures selected functional effects of potassium-mediated feedback while intentionally abstracting away detailed ion-channel biophysics to limit model complexity and retain the computational simplicity of the LIF framework. KvLIF primarily addresses intrinsic excitability regulation, whose implications for network-level homeostasis may depend on broader cellular and synaptic interactions \cite{refn1,refn2}.

\subsection{Dynamic Analysis of the KvLIF Model}
\begin{figure}[h]
\begin{center}
\centerline{\includegraphics[width=0.8\columnwidth]{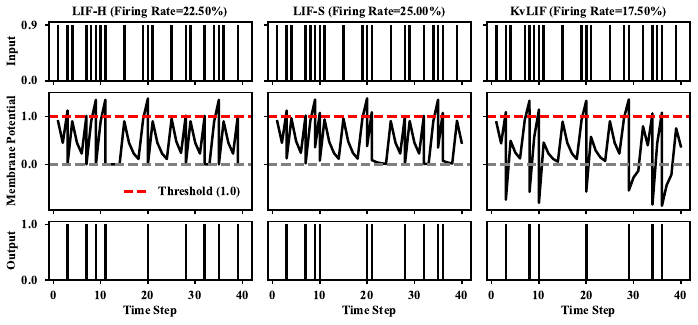}}
\caption{Comparison of dynamic responses between the KvLIF model and two conventional LIF models.}
\label{fig3}
\end{center}
\vskip -0.4in
\end{figure}

We visualize the dynamic response of the proposed KvLIF neuron to demonstrate its self-adaptation capabilities. The visualization of the variation in the internal variable \(K[t]\) is provided in Figure \ref{figr1} (Appendix \ref{a_r1}). As shown in Figure \ref{fig3}, all models are subjected to identical Poisson input spike trains over 40 time steps. 

\textbf{Regulation of Neuronal Excitability.}  Both conventional LIF models exhibit high firing rates, indicating a general tendency toward over-excitability. In contrast, the KvLIF neuron regulates its firing to a lower rate of 17.50\%, reflecting efficient self-adaptation that suppresses redundant spikes.

\textbf{Post-spike Membrane Dynamics.} In the soft-reset LIF, high post-spike residual potentials cause hypersensitivity to subsequent inputs (Figure \ref{fig3}, middle). The hard-reset LIF model forces the potential to the resting state, erasing the membrane potential history. The KvLIF neuron exhibits an afterhyperpolarization phase, where the potential drops significantly below the resting potential (Figure \ref{fig3}, right column). 
By incorporating the negative feedback mechanism, the KvLIF model inhibits excessive spiking, thereby enhancing the flexibility for encoding complex temporal patterns.

\begin{figure}[ht]
\centering
\subfigure[Gradient Norm Distribution \label{fig4a}]{
\centering
\includegraphics[width=0.4\columnwidth]{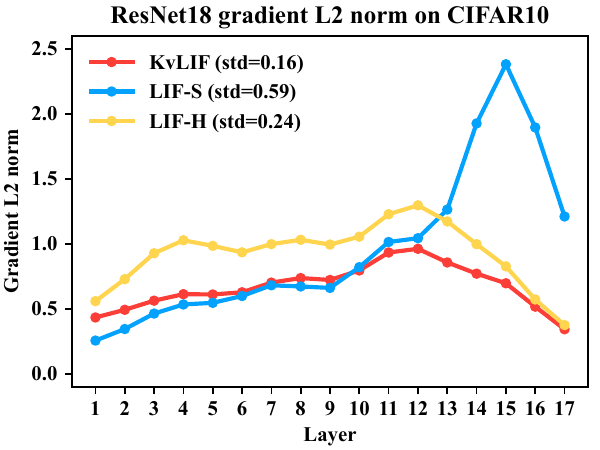}
}
\hspace{-5pt}
\subfigure[Adaptability Analysis \label{fig4b}]{
\centering
\includegraphics[width=0.4\columnwidth]{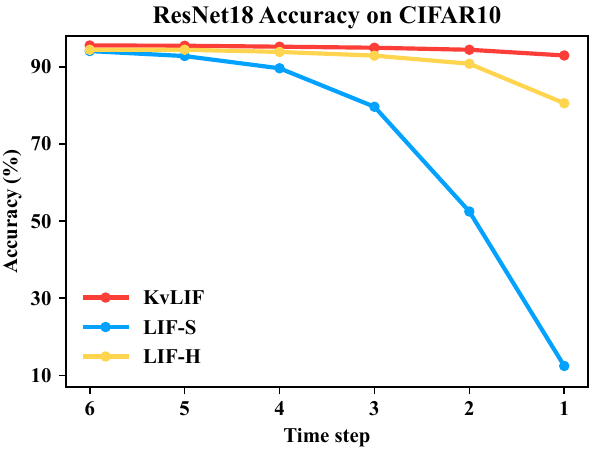}
}
\caption{Comparative performance of KvLIF and conventional LIF models. (a) Gradient L2 norm distribution. (b) Performance retention as inference time steps reduce.}
\vskip -0.1in
\label{fig4}
\end{figure}

\subsection{Gradient Stability and Adaptability Analysis of the KvLIF model}
The improved stability of KvLIF is shown in Figure \ref{fig4}(a). Layer-wise gradient L2 norms indicate that KvLIF enables more stable gradient propagation than conventional LIF models, as reflected by its lower standard deviation. Figure \ref{fig4}(b) further demonstrates its adaptability: although all models are trained with 6 time steps, KvLIF maintains over 90\% accuracy when tested with only one time step as inference time steps are progressively reduced. This result suggests that KvLIF supports adaptive encoding and efficient inference within a short temporal window, which is further validated on neuromorphic datasets in Section \ref{sec51}.

\section{Experiments}
\label{sec_exp}
We conduct a series of experiments to evaluate the proposed KvLIF neuron model, with all details provided in the Appendix \ref{a_b}.

\begin{table*}[t]
\centering
\caption{Performance comparison with existing methods on static image and neuromorphic datasets. The asterisk ($ \ast $) denotes implementations achieved solely by modifying neuron models within the open source code \cite{ref30}. \textsuperscript{\S}The CIFAR10-DVS test set contains 1,000 samples, so single-run accuracy changes in increments of 0.1 percentage points; the second decimal digit is therefore zero when percentages are reported to two decimal places.}
\label{tab1}
\vskip 0.1in
{\small
\begin{NiceTabularX}{0.99\textwidth}{>{\centering\arraybackslash}X|cccc}
\toprule
Dataset & Method & Network Architecture & Time step & Accuracy (\%) \\
\midrule
\multirow{9}{*}{CIFAR10} 
 & Dspike \cite{ref33} & Modified ResNet-18 & 4 / 6 & 93.66 / 94.05 \\
 & GLIF \cite{ref25} & ResNet-18 & 4 / 6 & 94.64 / 94.88 \\
 & SML \cite{ref35} & ResNet-18 & 4 / 6 & 95.01 / 95.12 \\
 & CLIF \cite{ref28} & ResNet-18 & 4 / 6 & 94.89 / 95.41 \\
 & ILIF \cite{ref29} & ResNet-18 & 4 / 6 & 95.24 / 95.49 \\
 & \textbf{KvLIF (Ours)} & \textbf{ResNet-18} & \textbf{4 / 6} & \textbf{95.28 / 95.55}\\
\cmidrule{2-5}
 & Temporal Pruning \cite{ref36} & VGG-16 & 5 & 93.90 \\
 & ILIF \cite{ref29} & VGG-16 & 6 & 94.25 \\
 & \textbf{KvLIF (Ours)} & \textbf{VGG-16} & \textbf{6} & \textbf{94.36} \\
\midrule
\multirow{9}{*}{CIFAR100} 
 & Dspike \cite{ref33} & Modified ResNet-18 & 4 / 6 & 73.35 / 74.24 \\
 & GLIF \cite{ref25} & ResNet-18 & 4 / 6 & 76.42 / 77.28 \\
 & SML \cite{ref35} & ResNet-18 & 4 / 6 & 77.36 / 78.00 \\
 & CLIF \cite{ref28} & ResNet-18 & 4 / 6 & 77.00 / 78.36 \\
 & ILIF \cite{ref29} & ResNet-18 & 4 / 6 & 77.43 / 78.51 \\
 & \textbf{KvLIF (Ours)} & \textbf{ResNet-18} & \textbf{4 / 6} & \textbf{77.76 / 79.30}\\
\cmidrule{2-5}
 & Temporal Pruning \cite{ref36} & VGG-16 & 5 & 71.58 \\
 & ILIF \cite{ref29} & VGG-16 & 6 & 75.25 \\
 & \textbf{KvLIF (Ours)} & \textbf{VGG-16} & \textbf{6} & \textbf{76.13} \\
 \midrule
 \multirow{4}{*}{Tiny ImageNet}
 & ASGL \cite{ref38}  & VGG-13 & 4 / 8 & 56.57 / 56.81 \\
 & CLIF \cite{ref28} & VGG-13 & 4 / 6 & 63.16 / 64.13 \\
 & CPT-SNN \cite{ref39} & VGG-13 & 4 / 8 & 63.39 / 64.78 \\
 & \textbf{KvLIF (Ours)} & \textbf{VGG-13} & \textbf{4 / 6} & \textbf{64.03 / 64.82}\\
\midrule
\multirow{10}{*}{CIFAR10-DVS\textsuperscript{\S}}
 & Dspike \cite{ref33} & ResNet-18 & 10 & 75.40 \\
 & PLIF \cite{ref23} & PLIF-Net & 20 & 74.80 \\
 & GLIF \cite{ref25} & 7B-wideNet & 16 & 78.10 \\
 & CLIF \cite{ref28} & VGG-11 & 16 & 79.00 \\
 & ILIF \cite{ref29} & VGG-11 & 10 & 78.60 \\
 & \textbf{KvLIF (Ours)} & \textbf{VGG-11} & \textbf{10} & \textbf{79.50}\\
\cmidrule{2-5}
 & PSN \cite{ref40} & VGGSNN & 10 & 85.90 \\
 & CLIF \cite{ref28} & VGGSNN & 10 & 86.10 \\
 & AR-LIF \cite{ref30} & VGGSNN & 16 & 87.90 \\
 & \textbf{KvLIF (Ours)\textsuperscript{*}} & \textbf{VGGSNN} & \textbf{10} & \textbf{88.40} \\
\midrule
\multirow{6}{*}{DVS-Gesture} 
 & PLIF \cite{ref23} & PLIF-Net & 20 & 97.57 \\
 & KLIF \cite{ref41} & Modified PLIF Net & 12 & 94.10 \\
 & CLIF \cite{ref28} & VGG-11 & 20 & 97.92 \\
 & AGMM \cite{ref42} & VGGSNN & 16 & 97.92 \\
 & ILIF \cite{ref29} & VGG-11 & 20 & 97.92 \\
 & \textbf{KvLIF (Ours)} & \textbf{VGG-11} & \textbf{10} & \textbf{98.26} \\
 \bottomrule
\end{NiceTabularX}
}
\vskip -0.1in
\end{table*}

\subsection{Performance Comparison}
\label{sec51}
We compare KvLIF with other competitive SNN methods, while detailed comparisons with conventional LIF models, SFA-based LIF models, and Artificial Neural Networks (ANNs) are provided in Appendices \ref{a_c} and \ref{a_r2}. To isolate the effect of the proposed neuron, we use only standard preprocessing and exclude advanced augmentation strategies such as AutoAugment and CutMix.

As shown in Table \ref{tab1}, KvLIF achieves consistently strong performance on both static image and neuromorphic benchmarks across different network architectures. On static image datasets, it obtains 95.55\% accuracy on CIFAR10 and 79.30\% on CIFAR100 with Spiking ResNet-18 using six time steps, and 64.82\% on Tiny ImageNet with Spiking VGG-13 using six time steps. Its advantage further extends to asynchronous spatiotemporal tasks: on CIFAR10-DVS \cite{ref43}, KvLIF achieves 79.50\% accuracy with Spiking VGG-11 using ten time steps and 88.40\% with VGGSNN, while reaching 98.26\% on DVS-Gesture \cite{ref44} with Spiking VGG-11 using ten time steps. We further evaluate KvLIF on three SNN Transformer architectures on CIFAR100, observing consistent accuracy gains over LIF (Appendix \ref{a_transformer}). Additional runs with multiple random seeds show consistent performance trends for KvLIF (Appendix \ref{a_r3}). In most cases, KvLIF outperforms competing methods with fewer time steps, demonstrating its effectiveness and efficiency in modeling both spatial and spatiotemporal information. Moreover, we analyze the firing rates and energy consumption of KvLIF (Appendix \ref{a_d} and \ref{a_e}).

To evaluate temporal generalization, we further test KvLIF on the SHD neuromorphic audio dataset, where it outperforms SFA-based baselines and shows improved capability in capturing long-range temporal dependencies from sparse neuromorphic audio sequences (Appendix \ref{a_r2}).

\begin{table}[h]
\centering
\caption{Robustness tests on CIFAR100 and DVS-Gesture under different noise levels.}
\label{tab3}
{
\setlength{\tabcolsep}{2pt}
\renewcommand{\arraystretch}{1.12}
\renewcommand{\tabularxcolumn}[1]{m{#1}}
\begin{tabularx}{\linewidth}{@{}>{\centering\arraybackslash}m{0.12\linewidth}*{10}{>{\centering\arraybackslash}X}@{}}
\toprule
& \multicolumn{5}{c}{CIFAR100} & \multicolumn{5}{c}{DVS-Gesture} \\
\cmidrule(lr){2-6}\cmidrule(l){7-11}
Neuron & 0.04 & 0.08 & 0.12 & 0.16 & 0.20 & 0.10 & 0.20 & 0.30 & 0.40 & 0.50 \\
\midrule
LIF-S & 65.58 & 53.96 & 40.81 & 29.74 & 21.57 & 90.63 & 84.03 & 77.78 & 69.10 & 66.32 \\
LIF-H & 71.60 & 61.23 & 44.35 & 30.57 & 21.33 & 77.78 & 68.40 & 60.07 & 55.56 & 52.08 \\
GLIF & 69.71 & 62.38 & 50.40 & 38.19 & 28.05 & 79.86 & 79.86 & 77.08 & 67.36 & 58.68 \\
CLIF & 75.57 & 66.71 & 60.71 & 40.83 & 29.93 & 89.58 & 82.64 & 75.69 & 63.89 & 57.29 \\
ILIF & 74.52 & 65.33 & 50.24 & 36.49 & 25.44 & 86.46 & 76.74 & 70.83 & 61.46 & 56.25 \\
KvLIF & \textbf{76.10} & \textbf{68.37} & \textbf{63.03} & \textbf{44.27} & \textbf{33.32} & \textbf{93.75} & \textbf{89.93} & \textbf{86.11} & \textbf{80.90} & \textbf{75.69} \\
\bottomrule
\end{tabularx}
}
\vskip -0.1in
\end{table}

\begin{table*}[h]
\centering
\caption{Ablation study of the proposed KvLIF neuron on different datasets.}
\label{tab5}
\vskip 0.1in
{\small
\setlength{\tabcolsep}{3.5pt}
\renewcommand{\arraystretch}{1.12}
\renewcommand{\tabularxcolumn}[1]{m{#1}}
\begin{tabularx}{0.96\textwidth}{@{}*{2}{>{\centering\arraybackslash}m{0.14\textwidth}}*{5}{>{\centering\arraybackslash}X}@{}}
\toprule
\shortstack{Synaptic\\shunting} &
\shortstack{Adaptive\\repolarization} &
CIFAR10 &
CIFAR100 &
\shortstack{Tiny\\ImageNet} &
\shortstack{CIFAR10\\-DVS} &
\shortstack{DVS-\\Gesture} \\
\midrule
\cmark & \cmark & 95.55 & 79.30 & 64.82 & 79.50 & 98.26 \\
\xmark & \cmark & 95.46 & 79.22 & 64.23 & 78.60 & 97.57 \\
\cmark & \xmark & 94.26 & 77.43 & 62.82 & 77.60 & 97.22 \\
\xmark & \xmark & 94.09 & 77.05 & 62.70 & 77.20 & 96.18 \\
\bottomrule
\end{tabularx}
}
\vskip -0.1in
\end{table*}

\subsection{Robustness Analysis} 
\label{sec52}
The adaptive regulation of neuronal excitability in KvLIF also enhances robustness to noise perturbations. 
For static image datasets, we use the Spiking ResNet-18 architecture and evaluate robustness by adding zero-mean Gaussian noise to the test set, with standard deviations ranging from 0.04 to 0.20.
For neuromorphic datasets, we employ the Spiking VGG-11 architecture. To preserve data sparsity, noise is added to 10\%-50\% of randomly selected pixels in the test set. Results for CIFAR100 and DVS-Gesture are presented in Table \ref{tab3}, with additional robustness evaluations on other datasets provided in the Appendix (Tables \ref{tab9} and \ref{tab10}, Appendix \ref{a_f}).

Across both static image and neuromorphic experiments, the classification accuracy of all models decreases as noise intensity increases. Nevertheless, KvLIF consistently achieves the highest accuracy under all tested noise levels, indicating stronger robustness to spatial perturbations in static images and event noise in neuromorphic data streams. In addition, KvLIF remains robust to missing temporal events, further demonstrating that its adaptive regulation enhances resilience to corrupted or incomplete spatiotemporal inputs (Tables \ref{tab11} and \ref{tab12}, Appendix \ref{a_g}).

\subsection{Ablation Study}
We analyze the relative contributions of the two adaptive components in the KvLIF model: the A-type current, which integrates membrane potential to shunt synaptic inputs, and the delayed rectifier current, which accumulates spiking history to regulate post-spike reset dynamics. Detailed quantitative results are presented in Table \ref{tab5}. Adaptive repolarization plays a more significant role in the improved performance of KvLIF, with its ablation causing a larger performance drop across all tested datasets. In comparison, synaptic shunting has a smaller impact.

\subsection{Hyperparameter Sensitivity Analysis}
We analyze the sensitivity of KvLIF to its key hyperparameters, $\alpha$, $\beta$, and $\gamma$. As shown in Table \ref{tab6} (Appendix \ref{a_b}), KvLIF requires limited per-dataset tuning: $\alpha$ and $\gamma$ are fixed at 0.8 and 0.05 across all five benchmarks, while $\beta$ only takes values of 0.1 or 0.3. Local sensitivity analysis around the default settings further shows that KvLIF remains stable under moderate hyperparameter perturbations. Moreover, $\gamma$ is fixed at 0.05 because it is the least sensitive coefficient, consistent with the ablation results in Table \ref{tab5}, where removing synaptic shunting causes a smaller performance drop than removing adaptive repolarization. The full results are provided in Appendix \ref{a_r4}.

\section{Conclusion}
In this study, we address key limitations of conventional LIF models, namely their restricted response range and susceptibility to noise, which hinder adaptability in complex environments. To overcome these challenges, KvLIF introduces a conductance state \(K[t]\) that integrates membrane-potential fluctuations and spiking history. Through this state, KvLIF dynamically modulates synaptic inputs before firing and adaptively adjusts the reset depth after firing. This coupled regulation of integration and reset expands the neuronal response range, suppresses redundant spikes, and improves temporal adaptability. Empirically, KvLIF achieves ANN-level accuracy on static benchmarks, suggesting that the proposed spiking neuron can narrow the conventional performance gap. Its advantages are more clearly reflected beyond accuracy: across static and neuromorphic datasets, KvLIF delivers stronger robustness, lower firing activity, and acceptable energy consumption. Overall, KvLIF highlights the value of potassium-inspired self-adaptation for building robust and efficient neuromorphic SNNs.

\section{Acknowledgments}

This work was supported by the National Key Research and Development Program of China (2023YFF1204200 to Y.Z.) and the National Natural Science Foundation of China (62476197 to Y.Z.).


\appendix

\section{Visualization of Feature Maps}
\label{a_a}
\begin{figure*}[th]
\begin{center}
\centerline{\includegraphics[width=0.8\columnwidth]{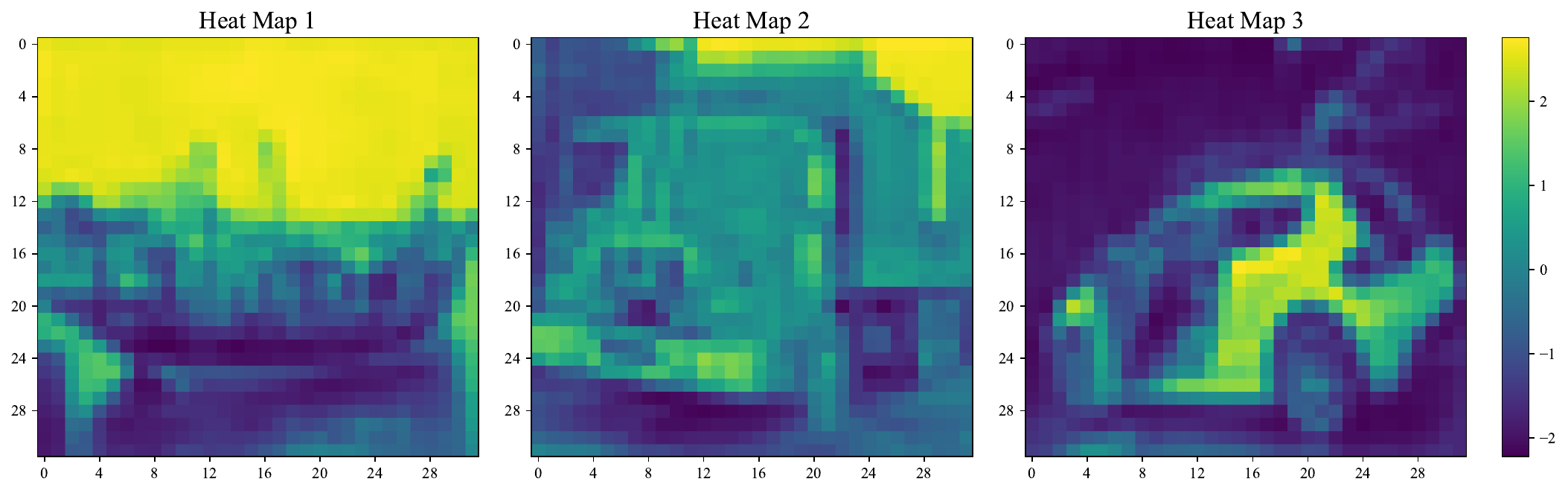}}
\caption{Heatmap visualization of the encoding layer responses to different input samples.}
\label{fig5}
\end{center}
\vskip -0.3in
\end{figure*}

We visualize the feature maps of the encoding layer in the LIF-based ResNet-18 on CIFAR10, as shown in Figure \ref{fig5}. Observations reveal that the input intensities received by LIF neurons significantly exceed 1.0, confirming that these neurons indeed process high-intensity signals.

\section{Core computation of the KvLIF model}
\label{a_r1n}
The pseudo-code for KvLIF (Algorithm \ref{alg1}) outlines the sequence of operations, illustrating how the potassium current state variable interacts with the membrane potential and preceding spiking history. 

Functioning as a dual-phase regulator, $K^l[t]$ modulates subthreshold depolarization through A-type potassium current-like mechanisms and controls repolarization depth via delayed rectifier potassium current-like mechanisms. This integrated regulation ensures that the integration and reset phases remain precisely coordinated at every time step.

\begin{algorithm}[h]
\caption{Core computation of the KvLIF model}
\label{alg1}
\textbf{Input:} Input current \(I\), time step \(t\), previous membrane potential \(U_{pre}\), previous state variable \(K_{pre}\).\\
\textbf{Hyperparameters:} Decay constant \(\lambda\), threshold \(V_{th}\), Potassium current decay factor \(\alpha\), membrane potential scaling factor \(\beta\), shunting effect scaling factor \(\gamma\).\\
\textbf{Output:} Spike \(S\)
\begin{algorithmic}[1]
\IF{\(t = 0\)}
\STATE Set \(U_{pre} \leftarrow 0\) and \(K_{pre} \leftarrow 0\) \hfill \(\triangleright\) initialization
\ENDIF
\STATE \(I \leftarrow I \odot (1 - \gamma \sigma(K_{pre}))\) \hfill \textcolor[RGB]{49, 126, 194} {\(\triangleright\) Shunting synaptic inputs}
\STATE \(U \leftarrow \lambda U_{pre} + I\) \hfill \(\triangleright\) leaky \& integrate
\STATE \(K_{init} \leftarrow \alpha K_{pre} + \beta U\) \hfill \(\triangleright\) Potassium currents
\STATE \(S \leftarrow H(U - V_{th})\) \hfill \(\triangleright\) fire
\STATE \(K \leftarrow K_{init} + S\) \hfill \(\triangleright\) Potassium currents surge
\STATE \(U \leftarrow U - S \odot V_{th}\) \hfill \(\triangleright\) reset
\STATE \(U \leftarrow U - S \odot \sigma(K)\) \hfill 
\textcolor[RGB]{130, 92, 166} {\(\triangleright\) Adaptive repolarization.}
\end{algorithmic}
\textbf{Return} \(S\)
\end{algorithm}

\section{Visualization of K variable}
\label{a_r1}
\begin{figure*}[th]
\begin{center}
\centerline{\includegraphics[width=0.4\columnwidth]{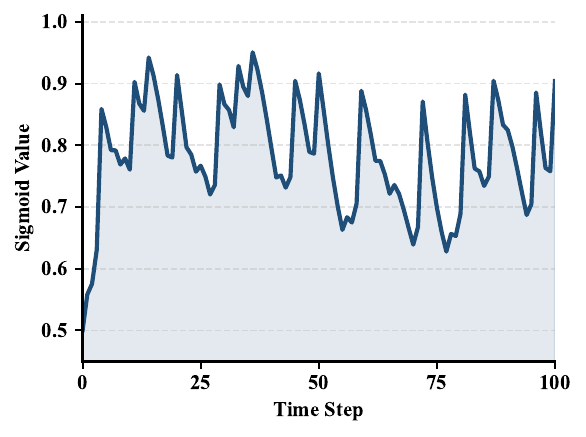}}
\caption{Temporal dynamics of the K variable.}
\label{figr1}
\end{center}
\vskip -0.3in
\end{figure*}

Under a constant input current of 0.8, we visualize the temporal variation of $ K[t] $ over 100 time steps. The results show that \(K[t]\) remains dynamically modulated throughout the simulation, rather than converging to a static state (Figure \ref{figr1}). This observation indicates that the dynamic behavior of $ K[t] $ retains activity-dependent regulatory properties.

\section{Detailed Experimental Settings and Dataset Preprocessing}
\label{a_b}
Experiments on all datasets except Tiny ImageNet are conducted on a single NVIDIA RTX 4090 GPU, while experiments on Tiny ImageNet are conducted on a single NVIDIA A800 GPU. We employ the Rectangle surrogate gradient function with $ \sigma = V_{th} = 1 $ and set the decay constant to 0.5. The random seed is fixed at 1000 for all runs. For the loss function, we utilize the TET method with a lambda value of 0.05 \cite{refa1}. Detailed training hyperparameters are listed in Table \ref{tab6}.

\begin{table}[htbp]
\centering
\caption{Training hyperparameter settings for various datasets.}
\label{tab6}
\vskip 0.1in
{\small
\setlength{\tabcolsep}{3pt}
\renewcommand{\arraystretch}{1.08}
\begin{tabular}{@{}
>{\centering\arraybackslash}m{0.17\linewidth}
>{\centering\arraybackslash}m{0.085\linewidth}
>{\centering\arraybackslash}m{0.075\linewidth}
>{\centering\arraybackslash}m{0.105\linewidth}
>{\centering\arraybackslash}m{0.11\linewidth}
>{\centering\arraybackslash}m{0.105\linewidth}
>{\centering\arraybackslash}m{0.065\linewidth}
>{\centering\arraybackslash}m{0.065\linewidth}
>{\centering\arraybackslash}m{0.07\linewidth}
@{}}
\toprule
\textbf{Dataset} &
\textbf{\shortstack{Batch\\size}} &
\textbf{Epoch} &
\textbf{\shortstack{Learning\\rate}} &
\textbf{Optimizer} &
\textbf{\shortstack{Weight\\decay}} &
$\boldsymbol{\alpha}$ &
$\boldsymbol{\beta}$ &
$\boldsymbol{\gamma}$ \\
\midrule
CIFAR10 & 128 & 200 & 0.1 & SGD & 5e-5 & 0.8 & 0.3 & 0.05 \\
CIFAR100 & 128 & 300 & 0.1 & SGD & 5e-4 & 0.8 & 0.1 & 0.05 \\
Tiny ImageNet & 256 & 300 & 0.1 & SGD & 5e-4 & 0.8 & 0.3 & 0.05 \\
CIFAR10-DVS & 128 & 200 & 0.05 & SGD & 5e-4 & 0.8 & 0.1 & 0.05 \\
DVS-Gesture & 16 & 150 & 5e-4 & Adam & 0.0 & 0.8 & 0.3 & 0.05 \\
\bottomrule
\end{tabular}
}
\end{table}

\textbf{CIFAR10 and CIFAR100.} These datasets each contain 60,000 color images (32 $\times$ 32 pixels), divided into 50,000 training and 10,000 testing samples across 10 and 100 classes, respectively. Data preprocessing includes random cropping with 4-pixel padding, random horizontal flipping, and Cutout. Image data are normalized to zero mean and unit variance. We employ the direct encoding method \cite{refa2}, where pixel values are repeatedly input into the model at each time step. The backbone network for these tasks is Spiking ResNet-18.

\textbf{Tiny ImageNet.} This dataset contains 100,000 color images (64 $\times$ 64 pixels) from 200 classes. Our preprocessing pipeline includes random cropping with 8-pixel padding and horizontal flipping. Direct encoding is utilized for input representation. The backbone network employed is Spiking VGG-13.

\textbf{CIFAR10-DVS.} This neuromorphic dataset comprises 10,000 event-driven samples derived from CIFAR-10, obtained by recording the original images using a Dynamic Vision Sensor (DVS) \cite{ref43}. Each sample features an original spatial resolution of 128 $\times$ 128 pixels. Each sample is processed using event-to-frame integration, with the spatial resolution at each time step resized to 48 $\times$ 48. This is followed by random cropping with 4-pixel padding to generate the final preprocessed samples. We adopt Spiking VGG-11 as the backbone network.

\textbf{DVS-Gesture.} This dataset captures 11 distinct hand gesture categories performed by 29 subjects under three different lighting conditions \cite{ref44}. It consists of 1,342 event-based samples (128 $\times$ 128 pixels), divided into 1,208 training and 134 testing samples. For preprocessing, we utilize the SpikingJelly framework to implement event-to-frame integration \cite{refa3}. The backbone network employed is Spiking VGG-11.

\section{Performance Comparison with Conventional LIF models and ANNs}
\label{a_c}
To evaluate the effectiveness of the proposed KvLIF neuron, we compare it with conventional LIF models on both static and neuromorphic datasets, and further include ANN counterparts with ReLU activations on CIFAR10 and CIFAR100, as summarized in Table \ref{tab7}. The experiments cover multiple network architectures and time steps, allowing us to assess whether KvLIF improves spiking dynamics while narrowing the performance gap between SNNs and ANNs.

Experimental results show that KvLIF consistently outperforms conventional LIF-S and LIF-H baselines across all evaluated datasets and architectures. On static image datasets, KvLIF achieves strong accuracy at both $T=4$ and $T=6$. Notably, on CIFAR10, KvLIF reaches 95.55\% at $T=6$, approaching the ReLU ANN counterpart with only a 0.16\% gap. On CIFAR100, KvLIF obtains 79.30\% at $T=6$, surpassing the ReLU ANN baseline by 0.99\%. On Tiny ImageNet, KvLIF also achieves the best performance among spiking models, reaching 64.82\% at $T=6$. For neuromorphic tasks, KvLIF demonstrates superior temporal processing ability, attaining 79.50\% on CIFAR10-DVS and 98.26\% on DVS-Gesture. These results indicate that the adaptive potassium-regulated dynamics not only improve over conventional LIF neurons, but also enable competitive or superior performance compared with ANN counterparts on static benchmarks where such comparisons are available.

\begin{table}[ht]
\centering
\caption{Performance comparison with conventional LIF models.}
\vskip 0.1in
\label{tab7}
\begin{tabular}{ccccc}
\hline
\textbf{Dataset} & \textbf{Architecture} & \textbf{Neuron} & \textbf{Time step} & \textbf{Accuracy (\%)} \\ \hline
\multirow{4}{*}{CIFAR10}
 & ResNet-18 & ReLU & 1 & 95.71 \\
 & ResNet-18 & LIF-S & 4 / 6 & 91.54 / 94.09 \\
 & ResNet-18 & LIF-H & 4 / 6 & 94.42 / 94.70 \\
 & \textbf{ResNet-18} & \textbf{KvLIF} & \textbf{4 / 6} & \textbf{95.28 / 95.55} \\ \hline
\multirow{4}{*}{CIFAR100}
 & ResNet-18 & ReLU & 1 & 78.31 \\
 & ResNet-18 & LIF-S & 4 / 6 & 69.24 / 77.05 \\
 & ResNet-18 & LIF-H & 4 / 6 & 74.64 / 77.17 \\
  & \textbf{ResNet-18} & \textbf{KvLIF} & \textbf{4 / 6} & \textbf{77.76 / 79.30} \\ \hline
\multirow{4}{*}{Tiny ImageNet}
 & VGG-13 & ReLU & 1 & 62.86 \\
 & VGG-13 & LIF-S & 4 / 6 & 60.72 / 62.70 \\
 & VGG-13 & LIF-H & 4 / 6 & 61.58 / 61.81 \\
 & \textbf{VGG-13} & \textbf{KvLIF} & \textbf{4 / 6} & \textbf{64.03 / 64.82} \\ \hline
 \multirow{3}{*}{CIFAR10-DVS} 
& VGG-11 & LIF-S & 10 & 77.20 \\
 & VGG-11 & LIF-H & 10 & 77.90 \\
 & \textbf{VGG-11} & \textbf{KvLIF} & \textbf{10} & \textbf{79.50} \\ \hline
\multirow{3}{*}{DVS-Gesture} 
& VGG-11 & LIF-S & 10 & 96.18 \\
 & VGG-11 & LIF-H & 10 & 96.88 \\
 & \textbf{VGG-11} & \textbf{KvLIF} & \textbf{10} & \textbf{98.26} \\ \hline
\end{tabular}
\end{table}

\section{Performance Comparison with LIF models incorporating SFA}
\label{a_r2}
KvLIF is compared with ALIF and adLIF across static image, neuromorphic vision, and neuromorphic audio datasets under different architectures and time steps. As shown in Table \ref{tabr1}, KvLIF consistently outperforms the baseline models on both static image and neuromorphic vision tasks, demonstrating its advantage in handling different visual input modalities.

The comparison is further extended to the SHD neuromorphic audio dataset in Table \ref{tabr2}. Under recurrent architectures with time steps no fewer than 250, KvLIF obtains 94.78$\,\pm\,$0.54\% with a one-layer RNN and 95.98$\,\pm\,$0.11\% with a two-layer RNN, both surpassing the corresponding adLIF models. These results indicate that KvLIF provides more effective temporal feature representation and exhibits strong generalization ability across different input modalities and network structures.

\begin{table}[ht]
\centering
\caption{Performance comparison with LIF models incorporating SFA.}
\vskip 0.1in
\label{tabr1}
\begin{tabular}{ccccc}
\hline
\textbf{Dataset} & \textbf{Architecture} & \textbf{Neuron} & \textbf{Time step} & \textbf{Accuracy (\%)} \\ \hline
\multirow{3}{*}{CIFAR10}
 & ResNet-18 & ALIF & 4 & 94.44 \\
 & ResNet-18 & adLIF & 4 & 94.41 \\
 & \textbf{ResNet-18} & \textbf{KvLIF} & \textbf{4} & \textbf{95.28} \\ \hline
\multirow{3}{*}{CIFAR100}
 & ResNet-18 & ALIF & 4 & 75.75 \\
 & ResNet-18 & adLIF & 4 & 72.99 \\
  & \textbf{ResNet-18} & \textbf{KvLIF} & \textbf{4} & \textbf{77.76} \\ \hline
 \multirow{3}{*}{CIFAR10-DVS} 
& VGG-11 & ALIF & 10 & 78.40 \\
 & VGG-11 & adLIF & 10 & 77.70 \\
 & \textbf{VGG-11} & \textbf{KvLIF} & \textbf{10} & \textbf{79.50} \\ \hline
\multirow{3}{*}{DVS-Gesture} 
& VGG-11 & ALIF & 10 & 97.22 \\
 & VGG-11 & adLIF & 10 & 97.92 \\
 & \textbf{VGG-11} & \textbf{KvLIF} & \textbf{10} & \textbf{98.26} \\ \hline
\end{tabular}
\end{table}

\begin{table}[ht]
\centering
\caption{Performance comparison on a neuromorphic audio dataset.}
\vskip 0.1in
\label{tabr2}
\begin{tabular}{ccccc}
\hline
\textbf{Dataset} & \textbf{Architecture} & \textbf{Neuron} & \textbf{Time step} & \textbf{Accuracy (\%)} \\ \hline
\multirow{5}{*}{SHD} 
& RNN & ALIF & 250 & 90.40 \\
 & RNN (1 layer) & adLIF & 250 & 94.59$\,\pm\,$0.27 \\
 & RNN (2 layers)& adLIF & 250 & 95.81$\,\pm\,$0.56 \\
 & \textbf{RNN (1 layer)} & \textbf{KvLIF} & \textbf{250} & \textbf{94.78$\,\pm\,$0.54} \\
 & \textbf{RNN (2 layers)} & \textbf{KvLIF} & \textbf{250} & \textbf{95.98$\,\pm\,$0.11} \\ \hline
\end{tabular}
\end{table}

\begin{figure}[ht]
    \centering
    \subfigure[CIFAR10. \label{figa1a}]{
        \centering
        \includegraphics[width=0.3\columnwidth]{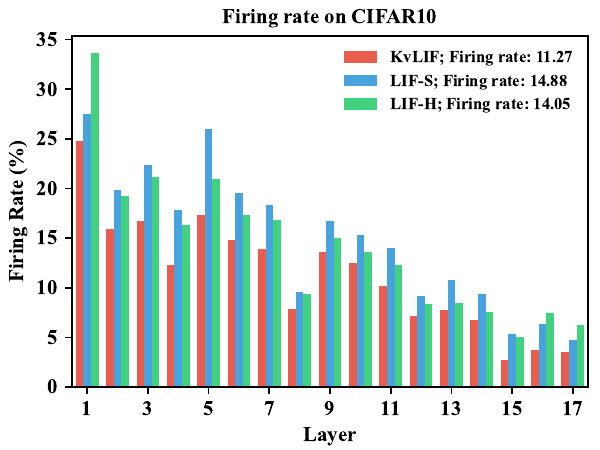}
    }
    \vspace{0.02\columnwidth}
    \subfigure[CIFAR100. \label{figa1b}]{
        \centering
        \includegraphics[width=0.3\columnwidth]{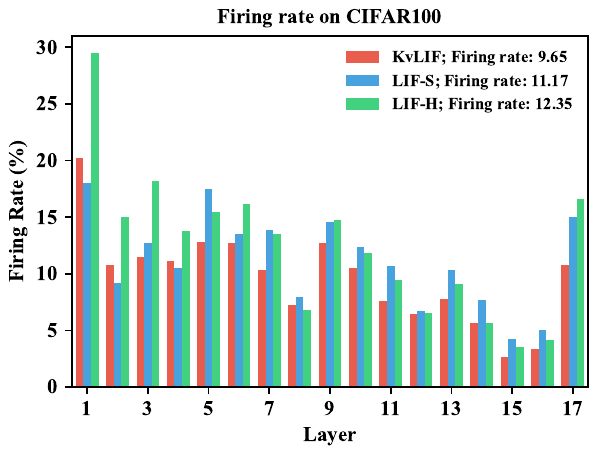}
    }
    \vspace{0.02\columnwidth}
    \subfigure[Tiny ImageNet. \label{figa1c}]{
        \centering
        \includegraphics[width=0.3\columnwidth]{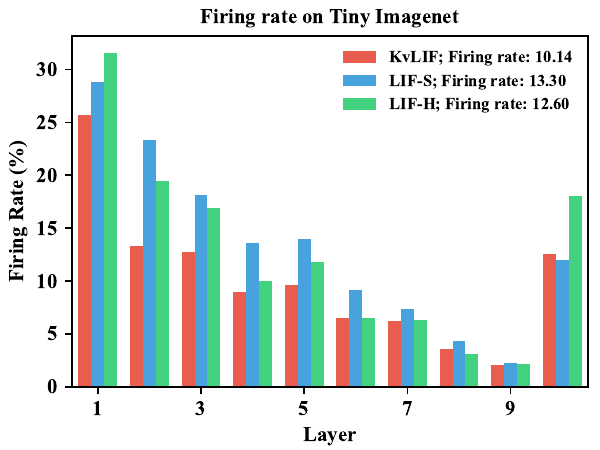}
    }
    \subfigure[CIFAR10-DVS \label{figa1d}]{
        \centering
        \includegraphics[width=0.3\columnwidth]{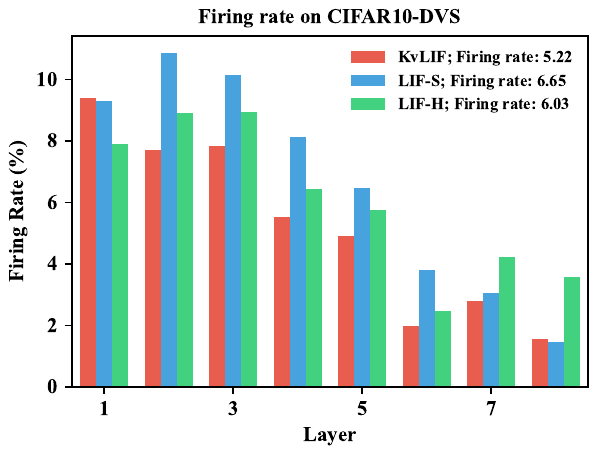}
    }
    \vspace{0.02\columnwidth}
    \subfigure[DVS-Gesture. \label{figa1e}]{
        \centering
        \includegraphics[width=0.3\columnwidth]{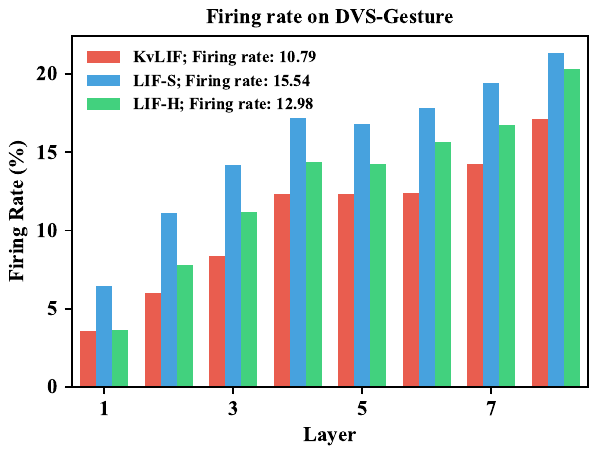}
    }
    \vspace{-0.1in}
    \caption{Firing rate comparison across different datasets.}
    \label{fig6}
\end{figure}

\section{Performance Evaluation on Transformer Architectures}
\label{a_transformer}
To demonstrate the broad applicability of KvLIF, we evaluated its performance on three SNN Transformer architectures, including Spikformer \cite{ref_spikformer}, Spike-Driven Transformer \cite{refa4}, and QKformer \cite{ref_qkformer}, using the CIFAR100 benchmark. As shown in Table \ref{tab:transformer_performance}, under identical training configurations and time steps, simply replacing standard LIF with KvLIF consistently yields accuracy gains across all Transformer backbones.

\begin{table}[ht]
\centering
\caption{Performance comparison on CIFAR100 with SNN Transformer architectures.}
\vskip 0.1in
\label{tab:transformer_performance}
\begin{tabular}{ccc}
\hline
\textbf{Architecture} & \textbf{Neuron} & \textbf{Accuracy (\%)} \\ \hline
\multirow{2}{*}{Spikformer} & LIF & 77.68 \\
 & \textbf{KvLIF} & \textbf{77.95} \\ \hline
\multirow{2}{*}{Spike-Driven Transformer} & LIF & 78.57 \\
 & \textbf{KvLIF} & \textbf{78.82} \\ \hline
\multirow{2}{*}{QKformer} & LIF & 80.57 \\
 & \textbf{KvLIF} & \textbf{81.43} \\ \hline
\end{tabular}
\end{table}

\section{Multi-seed evaluation across all datasets}
\label{a_r3}
To further evaluate the reliability of the proposed KvLIF model, we conduct multi-seed experiments across all datasets. The results are reported as the mean and standard deviation over three random seeds. As shown in Table \ref{tabr3}, KvLIF consistently maintains strong performance across different datasets, architectures, and input modalities. The small performance variations indicate that the improvements brought by KvLIF are not dependent on a specific random initialization. These results further confirm the consistent performance across the evaluated random initializations.

\begin{table}[ht]
\centering
\caption{Multi-seed results across all datasets.}
\vskip 0.1in
\label{tabr3}
\begin{tabular}{ccccc}
\hline
\textbf{Dataset} & \textbf{Architecture} & \textbf{Neuron} & \textbf{Time step} & \textbf{Accuracy (\%)} \\ \hline
\multirow{1}{*}{CIFAR10}
 & ResNet-18 & KvLIF & 4 & 95.23$\,\pm\,$0.04 \\ \hline
\multirow{1}{*}{CIFAR100}
 & ResNet-18 & KvLIF & 4 & 77.85$\,\pm\,$0.15 \\ \hline
\multirow{1}{*}{Tiny ImageNet}
 & VGG-13 & KvLIF & 4 & 64.07$\,\pm\,$0.06 \\ \hline
 \multirow{1}{*}{CIFAR10-DVS} 
& VGG-11 & KvLIF & 10 & 79.60$\,\pm\,$0.14 \\ \hline
\multirow{1}{*}{DVS-Gesture} 
& VGG-11 & KvLIF & 10 & 98.26$\,\pm\,$0.00 \\ \hline
\end{tabular}
\end{table}

\section{Firing Rate Analysis}
\label{a_d}
We calculate the layer-wise firing rates of KvLIF compared to conventional LIF models across five datasets, as shown in Figure \ref{fig6}. The results demonstrate that KvLIF consistently achieves lower firing rates on both static image datasets (CIFAR10, CIFAR100, Tiny ImageNet) and neuromorphic datasets (CIFAR10-DVS, DVS-Gesture). This significant reduction in spiking activity is attributed to the intrinsic adaptive mechanism of the KvLIF neuron. By effectively suppressing the generation of redundant spikes, KvLIF ensures that only task-relevant information is transmitted, thereby maximizing computational sparsity.

\section{Energy Consumption Analysis}
\label{a_e}
To evaluate the energy efficiency of the proposed model, we analyze the total energy consumption based on Synaptic Operations (SOP) \cite{refa4}. The total energy consumption is calculated by summing the accumulation ($AC$) and multiply-accumulate ($MAC$) operations, weighted by their respective unit energy costs. The formulation is given by:
\begin{equation}
SOP = E_{AC} \cdot N_{AC} + E_{MAC} \cdot N_{MAC},
\end{equation}
where $E_{AC}$ and $E_{MAC}$ denote the energy cost per operation, and $N_{AC}$ and $N_{MAC}$ represent the total number of $AC$ and $MAC$ operations, respectively. Following established energy models in neuromorphic computing, we set $E_{AC}$ to $0.9$ pJ for one 32-bit floating-point $AC$ and $E_{MAC}$ to $4.6$ pJ for one 32-bit floating-point MAC operation \cite{refa5}.

This efficiency analysis is based on $AC$/$MAC$ accounting and provides a hardware-agnostic reference for computational complexity, rather than a precise estimate of deployment cost. In particular, it does not include the overhead of sigmoid nonlinearities or the storage and memory read/write costs introduced by the additional potassium state variable. Thus, the reported complexity should be viewed as a basic comparison across models rather than a complete measure of runtime overhead on specific hardware platforms.

In our implementation, the total cost of each layer $l$ is divided into two parts: (1) spike-driven synaptic accumulations and (2) neuron-state updates. We define $C_{\mathrm{dense}}^{(l)}$ as the ANN-equivalent dense operation count for layer $l$, i.e., $C_{\mathrm{out}}H_{\mathrm{out}}W_{\mathrm{out}}C_{\mathrm{in}}k_hk_w$ for a convolutional layer, and $N_{\mathrm{in}}N_{\mathrm{out}}$ for a fully connected layer. For SNNs, this dense computation is converted into synaptic $AC$s according to
\begin{equation}
N_{\mathrm{AC}}^{(l)} = C_{\mathrm{dense}}^{(l)} \cdot fr_{\mathrm{in}}^{(l)} \cdot T,
\end{equation}
where $fr_{\mathrm{in}}^{(l)}$ is the input firing rate of layer $l$ and $T$ is the simulation length. The $MAC$ term is used separately for neuron-state updates. KvLIF has a larger neuron-level $MAC$ term because, beyond standard membrane integration and reset, it additionally updates the potassium state, applies shunting modulation to the input current, and performs adaptive repolarization.

The network-level $N_{\mathrm{AC}}$ and $N_{\mathrm{MAC}}$ are obtained by summing the corresponding terms over all layers. As shown in Table \ref{tab8}, although KvLIF introduces additional local adaptive computations for potassium current dynamics and thus increases the neuron-level $MAC$ cost, its lower firing rate reduces spike-driven synaptic $AC$s across layers. This reduction effectively offsets the extra computation, allowing KvLIF to maintain comparable or even lower overall energy consumption than conventional LIF models. These results indicate that incorporating potassium channel mechanisms can improve the accuracy of SNNs while preserving their energy efficiency.

\begin{table}[ht]
\centering
\caption{The energy consumption of synaptic operation on different datasets.}
\vskip 0.1in
\label{tab8}
\begin{tabular}{ccccccc}
\hline
\textbf{Dataset} & \textbf{Architecture} & \textbf{Neuron} & \textbf{T} & \textbf{ACs (M)} & \textbf{MACs (M)} & \textbf{\begin{tabular}[c]{@{}c@{}}SOP Energy ($\mu$J)\end{tabular}} \\ \hline
\multirow{4}{*}{CIFAR10} & ResNet-18 & ReLU & 1 & 0.00 & 549.13 & 2526.00 \\
 & ResNet-18 & LIF-S & 6 & 512.82 & 3.34 & 476.91 \\
 & ResNet-18 & LIF-H & 6 & 487.93 & 3.34 & 454.51 \\
 & ResNet-18 & KvLIF & 6 & 393.16 & 13.37 & 415.34 \\ \hline
\multirow{4}{*}{CIFAR100} & ResNet-18 & ReLU & 1 & 0.00 & 549.18 & 2526.22 \\
 & ResNet-18 & LIF-S & 6 & 359.34 & 3.34 & 338.78 \\
 & ResNet-18 & LIF-H & 6 & 404.70 & 3.34 & 379.61 \\
 & ResNet-18 & KvLIF & 6 & 320.60 & 13.37 & 350.05 \\ \hline
\multirow{4}{*}{Tiny ImageNet} & VGG-13 & ReLU & 1 & 0.00 & 913.46 & 4201.90 \\
 & VGG-13 & LIF-S & 6 & 885.85 & 6.00 & 824.86 \\
 & VGG-13 & LIF-H & 6 & 822.80 & 6.00 & 768.11 \\
 & VGG-13 & KvLIF & 6 & 679.14 & 23.99 & 721.58 \\ \hline
\multirow{3}{*}{DVS-Gesture} & VGG-11 & LIF-S & 10 & 3674.48 & 24.25 & 3418.58 \\
 & VGG-11 & LIF-H & 10 & 2992.54 & 24.25 & 2804.83 \\
 & VGG-11 & KvLIF & 10 & 2503.10 & 96.99 & 2698.96 \\ \hline
\multirow{3}{*}{CIFAR10-DVS} & VGG-11 & LIF-S & 10 & 302.70 & 3.41 & 288.15 \\
 & VGG-11 & LIF-H & 10 & 264.66 & 3.41 & 253.88 \\
 & VGG-11 & KvLIF & 10 & 241.49 & 13.64 & 280.12 \\ \hline
\end{tabular}
\end{table}

To account for costs beyond the operation-based estimate above, we further include memory reads and writes, synaptic and neuronal operations, and data addressing, following prior work \cite{ref28,refa6}, with an estimated cost of 12.35~pJ per sigmoid evaluation \cite{refa7}. Although the broader accounting in Table \ref{tab:extended_energy} indicates higher total energy consumption for KvLIF than for LIF, this additional cost accompanies consistent gains in accuracy and robustness. Moreover, KvLIF retains a substantial energy advantage over the corresponding ReLU ANN on static datasets, supporting a favorable performance--energy trade-off.

\begin{table}[ht]
\centering
\caption{Extended energy estimates including memory reads and writes, synaptic and neuronal computation, and data addressing.}
\vskip 0.1in
\label{tab:extended_energy}
\begin{tabular}{cccccc}
\hline
\textbf{Dataset} & \textbf{Neuron} & \begin{tabular}[c]{@{}c@{}}\textbf{Mem. Read}\\\textbf{\& Write (mJ)}\end{tabular} & \begin{tabular}[c]{@{}c@{}}\textbf{Synaptic \&}\\\textbf{Neuron Op. (mJ)}\end{tabular} & \textbf{Addr. (mJ)} & \textbf{Total (mJ)} \\
\hline
\multirow{3}{*}{CIFAR10}
 & ReLU  & 109.9045 & 1.7573 & 0.0001 & 111.6619 \\
 & LIF   & 34.8653  & 0.0190 & 0.0123 & 34.8966  \\
 & KvLIF & 61.5563  & 0.1211 & 0.0185 & 61.6960  \\
\hline
\multirow{3}{*}{CIFAR100}
 & ReLU  & 109.9091 & 1.7574 & 0.0001 & 111.6667 \\
 & LIF   & 25.3054  & 0.0170 & 0.0088 & 25.3312  \\
 & KvLIF & 50.9449  & 0.1203 & 0.0153 & 51.0804  \\
\hline
\multirow{3}{*}{Tiny ImageNet}
 & ReLU  & 182.8268 & 2.9232 & 0.0002 & 185.7501 \\
 & LIF   & 44.5320  & 0.0446 & 0.0157 & 44.5923  \\
 & KvLIF & 79.8119  & 0.2282 & 0.0241 & 80.0642  \\
\hline
\multirow{2}{*}{CIFAR10-DVS}
 & LIF   & 9.1751   & 0.0234 & 0.0031 & 9.2016   \\
 & KvLIF & 17.1263  & 0.1295 & 0.0050 & 17.2608  \\
\hline
\multirow{2}{*}{DVS-Gesture}
 & LIF   & 110.8881 & 0.1168 & 0.0377 & 111.0425 \\
 & KvLIF & 176.7999 & 0.8593 & 0.0514 & 177.7106 \\
\hline
\end{tabular}
\end{table}

\section{Additional Robustness Test}
\label{a_f}
We conduct additional robustness tests on the static image dataset (CIFAR-10) and the neuromorphic dataset (CIFAR10-DVS). The noise settings follow the protocols described in Section \ref{sec52} of the main text, and the results are summarized in Tables \ref{tab9} and \ref{tab10}.

In the static image experiments, accuracy naturally degrades for all models as noise intensity increases. The KvLIF model consistently achieves higher accuracy than other neuron models across almost all noise levels. The only exception is at the 0.08 noise level, where it is slightly surpassed by others. Similarly, in the neuromorphic experiments, while all models exhibit reduced accuracy with increasing noise proportions, KvLIF maintains the highest accuracy at all levels, demonstrating superior resilience to event noise.

\begin{table}[H]
\centering
\caption{Robustness test on CIFAR10 under different noise levels.}
\vskip 0.1in
\label{tab9}
\begin{tabular}{ccccccccc}
\toprule
Neuron & 0.04 & 0.08 & 0.12 & 0.16 & 0.20 \\
\midrule
LIF-S & 90.65 & 86.74 & 79.91 & 70.96 & 60.75 \\
LIF-H & 93.72 & 90.87 & 86.38 & 79.59 & 71.82 \\
GLIF & 93.64 & 91.48 & 87.92 & 82.05 & 74.97 \\
CLIF & 94.44 & 92.46 & 88.40 & 82.69 & 74.98 \\
ILIF & 94.31 & \textbf{92.96} & 88.77 & 83.63 & 76.61 \\
KvLIF & \textbf{94.84} & 92.73 & \textbf{89.12} & \textbf{84.12} & \textbf{77.02} \\
\bottomrule
\end{tabular}
\end{table}

\begin{table}[H]
\centering
\caption{Robustness test on CIFAR10-DVS under different noise levels.}
\vskip 0.1in
\label{tab10}
\begin{tabular}{ccccccccc}
\toprule
Neuron & 0.10 & 0.20 & 0.30 & 0.40 & 0.50 \\
\midrule
LIF-S & 75.20 & 73.80 & 69.30 & 65.30 & 58.80 \\
LIF-H & 75.90 & 74.70 & 69.20 & 66.30 & 61.20 \\
GLIF & 70.20 & 61.40 & 51.00 & 40.60 & 31.30 \\
CLIF & 75.40 & 73.20 & 69.60 & 66.10 & 60.30 \\
ILIF & 75.80 & 75.70 & 72.80 & 70.00 & 64.40 \\
KvLIF & \textbf{77.10} & \textbf{75.90} & \textbf{74.40} & \textbf{72.10} & \textbf{69.00} \\
\bottomrule
\end{tabular}
\end{table}

\section{Temporal Robustness Analysis}
To verify the temporal robustness of KvLIF, we introduce temporal noise by randomly masking out a certain percentage of event time steps in the test sets. These experiments are conducted on both the DVS-Gesture and CIFAR10-DVS datasets, with drop rates ranging from 0.1 to 0.5.

\label{a_g}
\begin{table}[H]
\centering
\caption{Temporal Robustness test on DVS-Gesture under different noise levels.}
\vskip 0.1in
\label{tab11}
\begin{tabular}{ccccccccc}
\toprule
Neuron & 0.10 & 0.20 & 0.30 & 0.40 & 0.50 \\
\midrule
LIF-S & 96.53 & 95.83 & 94.10 & 92.36 & 89.93 \\
LIF-H & 95.83 & 95.14 & 90.28 & 78.47 & 60.07 \\
GLIF & 95.83 & 96.18 & 96.18 & 94.79 & 93.06 \\
CLIF & 96.18 & 95.49 & 94.44 & 93.06 & 90.97 \\
ILIF & 96.86 & 96.53 & 94.44 & 92.01 & 88.54 \\
KvLIF & \textbf{97.22} & \textbf{96.53} & \textbf{96.53} & \textbf{95.83} & \textbf{94.10} \\
\bottomrule
\end{tabular}
\end{table}

While the accuracy of all models naturally declines as noise levels increase, the KvLIF model demonstrates stronger temporal robustness. Specifically, on the DVS-Gesture dataset, KvLIF maintains the highest accuracy across all drop rates (Table \ref{tab11}). Similarly, on CIFAR10-DVS, it outperforms competing models at all noise levels, with the sole exception of the 0.4 drop rate  (Table \ref{tab12}).

\begin{table}[H]
\centering
\caption{Temporal Robustness test on CIFAR10-DVS under different noise levels.}
\vskip 0.1in
\label{tab12}
\begin{tabular}{ccccccccc}
\toprule
Neuron & 0.10 & 0.20 & 0.30 & 0.40 & 0.50 \\
\midrule
LIF-S & 75.40 & 74.10 & 71.50 & 66.60 & 55.10 \\
LIF-H & 77.40 & 76.00 & 74.00 & 71.40 & 61.60 \\
GLIF & 74.10 & 69.70 & 63.80 & 55.50 & 43.70 \\
CLIF & 77.90 & 76.30 & 74.10 & 68.50 & 57.70 \\
ILIF & 78.10 & 76.20 & 74.60 & \textbf{71.60} & 63.80 \\
KvLIF & \textbf{79.00} & \textbf{77.90} & \textbf{74.70} & 71.10 & \textbf{65.00} \\
\bottomrule
\end{tabular}
\end{table}

\section{Hyperparameter Sensitivity Analysis Results}
\label{a_r4}
To further evaluate the robustness of KvLIF to hyperparameter choices, we conduct a local sensitivity analysis around the default settings of $\alpha$, $\beta$, and $\gamma$. Specifically, we vary $\alpha$ and $\beta$ within a moderate range while keeping $\gamma$ fixed at 0.05, and additionally test different values of $\gamma$ across multiple datasets (Tables \ref{tabr4}, \ref{tabr5}, \ref{tabr6}, and \ref{tabr7}). The results show that KvLIF maintains stable performance on CIFAR-10, CIFAR10-DVS, and DVS-Gesture under moderate changes of $\alpha$ and $\beta$, indicating that the model is not highly dependent on precise hyperparameter tuning. CIFAR-100 exhibits relatively higher sensitivity, especially with respect to $\beta$, but the performance remains competitive across a broad range of parameter settings. 

For $\gamma$, the results are consistently stable across all tested datasets (Table \ref{tabr8}), supporting our choice of fixing $\gamma$=0.05 in the main experiments. Overall, these results demonstrate that KvLIF achieves robust performance with only mild adjustment of $\beta$ for datasets with different input statistics.

\begin{table}[H]
\centering
\caption{Hyperparameter sensitivity analysis on CIFAR10 ($\gamma$ is fixed at 0.05)}
\vskip 0.1in
\label{tabr4}
\begin{tabular}{c|ccc}
\hline
$\alpha \backslash \beta$ & 0.2 & 0.3 & 0.4 \\
\hline
0.7 & 95.26\% & 95.00\% & 95.13\% \\
0.8 & 95.12\% & 95.28\% & 95.03\% \\
0.9 & 95.21\% & 95.07\% & 95.13\% \\
\hline
\end{tabular}
\end{table}

\begin{table}[H]
\centering
\caption{Hyperparameter sensitivity analysis on CIFAR100 ($\gamma$ is fixed at 0.05)}
\vskip 0.1in
\label{tabr5}
\begin{tabular}{c|ccc}
\hline
$\alpha \backslash \beta$ & 0.2 & 0.3 & 0.4 \\
\hline
0.7 & 77.33\% & 77.06\% & 77.02\% \\
0.8 & 78.03\% & 77.76\% & 76.75\% \\
0.9 & 77.66\% & 75.46\% & 77.43\% \\
\hline
\end{tabular}
\end{table}

\begin{table}[H]
\centering
\caption{Hyperparameter sensitivity analysis on CIFAR10-DVS ($\gamma$ is fixed at 0.05)}
\vskip 0.1in
\label{tabr6}
\begin{tabular}{c|ccc}
\hline
$\alpha \backslash \beta$ & 0.2 & 0.3 & 0.4 \\
\hline
0.7 & 78.70\% & 79.20\% & 78.90\% \\
0.8 & 78.90\% & 79.50\% & 79.10\% \\
0.9 & 79.00\% & 78.70\% & 78.60\% \\
\hline
\end{tabular}
\end{table}

\begin{table}[H]
\centering
\caption{Hyperparameter sensitivity analysis on DVS-Gesture ($\gamma$ is fixed at 0.05)}
\vskip 0.1in
\label{tabr7}
\begin{tabular}{c|ccc}
\hline
$\alpha \backslash \beta$ & 0.2 & 0.3 & 0.4 \\
\hline
0.7 & 97.92\% & 97.57\% & 97.92\% \\
0.8 & 97.22\% & 98.26\% & 97.57\% \\
0.9 & 97.57\% & 97.57\% & 97.57\% \\
\hline
\end{tabular}
\end{table}

\begin{table}[H]
\centering
\caption{Hyperparameter sensitivity analysis results for $\gamma$.}
\vskip 0.05in
\label{tabr8}
\begin{tabular}{c|ccc}
\hline
Dataset & 0.01 & 0.05 & 0.1 \\
\hline
CIFAR10 & 94.96\% & 95.28\% & 95.07\% \\
CIFAR100 & 77.63\% & 77.76\% & 76.19\% \\
CIFAR10-DVS & 78.90\% & 79.50\% & 78.90\% \\
DVS-Gesture & 97.92\% & 98.26\% & 97.57\% \\
\hline
\end{tabular}
\end{table}

\section{Limitations and Future Work}
\label{lim_fut}

While KvLIF demonstrates improvements in accuracy, robustness, and temporal representation, several limitations remain. First, our energy analysis includes both $AC$/$MAC$-based estimates and extended accounting for memory reads and writes, data addressing, and nonlinear computations (Appendix \ref{a_e}). The extended estimates indicate that KvLIF incurs higher total energy consumption than the evaluated LIF baseline, despite its reduced spiking activity, while remaining below the corresponding ReLU ANNs on static datasets. These results highlight a trade-off between improved performance and the overhead of adaptive neuronal dynamics. However, these estimates rely on assumed operation and memory-access costs and have not been validated through measurements on neuromorphic hardware. Future work should evaluate energy consumption and latency on specific platforms and investigate efficient state storage and nonlinear-function implementations to reduce the overhead of potassium-mediated adaptation.

Second, our evaluation focuses on classification tasks involving static images, neuromorphic vision, and neuromorphic audio. Although these benchmarks support the applicability of KvLIF across multiple input modalities and network architectures, its effectiveness beyond classification remains unverified. Future work could examine temporal detection, event-based tracking, and sequential decision-making to assess whether the proposed adaptive dynamics provide similar benefits under continuous inputs and task-specific temporal constraints.

\section{Impact Statement}
\label{imp_sta}
This work bridges biophysical principles and efficient model design, demonstrating how biologically inspired self-regulation can create more robust and adaptable SNNs. By improving noise resilience and dynamic range, it directly contributes to the development of reliable and energy-efficient neuromorphic computing for real-world applications like edge AI and autonomous systems.

\section{Code Availability}
The code for the KvLIF neuron model and the accompanying experiments is publicly available at https://github.com/ZubinYang/KvLIF.


\newpage
\input{checklist.tex}

\end{document}

%% file: checklist.tex
\section*{NeurIPS Paper Checklist}

\begin{enumerate}

\item {\bf Claims}
    \item[] Question: Do the main claims made in the abstract and introduction accurately reflect the paper's contributions and scope?
    \item[] Answer: \answerYes{} 
    \item[] Justification: We have summarized the main contributions and scope in the abstract and introduction sections.

\item {\bf Limitations}
    \item[] Question: Does the paper discuss the limitations of the work performed by the authors?
    \item[] Answer: \answerYes{} 
    \item[] Justification: We point out the limitations in Appendix \ref{lim_fut}.

\item {\bf Theory assumptions and proofs}
    \item[] Question: For each theoretical result, does the paper provide the full set of assumptions and a complete (and correct) proof?
    \item[] Answer: \answerYes{} 
    \item[] Justification: We present the assumptions and provide a complete proof in Section \ref{sec_met}

    \item {\bf Experimental result reproducibility}
    \item[] Question: Does the paper fully disclose all the information needed to reproduce the main experimental results of the paper to the extent that it affects the main claims and/or conclusions of the paper (regardless of whether the code and data are provided or not)?
    \item[] Answer: \answerYes{} 
    \item[] Justification: We have listed the training details in Appendix \ref{a_b}, from which people can reproduce the main experimental results based on it. Our code is publicly available on GitHub.

\item {\bf Open access to data and code}
    \item[] Question: Does the paper provide open access to the data and code, with sufficient instructions to faithfully reproduce the main experimental results, as described in supplemental material?
    \item[] Answer: \answerYes{} 
    \item[] Justification: We provide code in the supplementary materials. The open-source datasets are listed in Appendix \ref{a_b}.

\item {\bf Experimental setting/details}
    \item[] Question: Does the paper specify all the training and test details (e.g., data splits, hyperparameters, how they were chosen, type of optimizer) necessary to understand the results?
    \item[] Answer: \answerYes{} 
    \item[] Justification: We discuss the experimental setup and training details in Section \ref{sec_exp} and Appendix \ref{a_b}.

\item {\bf Experiment statistical significance}
    \item[] Question: Does the paper report error bars suitably and correctly defined or other appropriate information about the statistical significance of the experiments?
    \item[] Answer: \answerYes{} 
    \item[] Justification: We report the multi-seed evaluation across all datasets.

\item {\bf Experiments compute resources}
    \item[] Question: For each experiment, does the paper provide sufficient information on the computer resources (type of compute workers, memory, time of execution) needed to reproduce the experiments?
    \item[] Answer: \answerYes{} 
    \item[] Justification: We summarize our computational resources in Appendix \ref{a_b}.
    
\item {\bf Code of ethics}
    \item[] Question: Does the research conducted in the paper conform, in every respect, with the NeurIPS Code of Ethics \url{https://neurips.cc/public/EthicsGuidelines}?
    \item[] Answer: \answerYes{} 
    \item[] Justification: We have read the Code of Ethics and make sure to preserve anonymity.

\item {\bf Broader impacts}
    \item[] Question: Does the paper discuss both potential positive societal impacts and negative societal impacts of the work performed?
    \item[] Answer: \answerYes{} 
    \item[] Justification: We have discussed potential societal impacts in Appendix \ref{imp_sta}.
    
\item {\bf Safeguards}
    \item[] Question: Does the paper describe safeguards that have been put in place for responsible release of data or models that have a high risk for misuse (e.g., pre-trained language models, image generators, or scraped datasets)?
    \item[] Answer: \answerNA{} 
    \item[] Justification: The paper poses no such risks.

\item {\bf Licenses for existing assets}
    \item[] Question: Are the creators or original owners of assets (e.g., code, data, models), used in the paper, properly credited and are the license and terms of use explicitly mentioned and properly respected?
    \item[] Answer: \answerYes{} 
    \item[] Justification: We have used open-source code and datasets, and have appropriately cited them.

\item {\bf New assets}
    \item[] Question: Are new assets introduced in the paper well documented and is the documentation provided alongside the assets?
    \item[] Answer: \answerNA{} 
    \item[] Justification: This work does not release new assets.

\item {\bf Crowdsourcing and research with human subjects}
    \item[] Question: For crowdsourcing experiments and research with human subjects, does the paper include the full text of instructions given to participants and screenshots, if applicable, as well as details about compensation (if any)? 
    \item[] Answer: \answerNA{} 
    \item[] Justification: Our paper does not involve crowdsourcing nor research with human subjects.

\item {\bf Institutional review board (IRB) approvals or equivalent for research with human subjects}
    \item[] Question: Does the paper describe potential risks incurred by study participants, whether such risks were disclosed to the subjects, and whether Institutional Review Board (IRB) approvals (or an equivalent approval/review based on the requirements of your country or institution) were obtained?
    \item[] Answer: \answerNA{} 
    \item[] Justification: Our paper does not involve crowdsourcing nor research with human subjects.

\item {\bf Declaration of LLM usage}
    \item[] Question: Does the paper describe the usage of LLMs if it is an important, original, or non-standard component of the core methods in this research? Note that if the LLM is used only for writing, editing, or formatting purposes and does \emph{not} impact the core methodology, scientific rigor, or originality of the research, declaration is not required.
    \item[] Answer: \answerNA{} 
    \item[] Justification: The core method development in this research does not involve LLMs as any important, original, or non-standard components.

\end{enumerate}